\documentclass[10pt,twocolumn,letterpaper]{article}

\usepackage{cvpr}

\usepackage{multirow}

\makeatletter
\renewcommand{\paragraph}{\@startsection{paragraph}{4}{\z@}%
  {0.35ex \@plus .1ex \@minus .1ex}%
  {-1em}%
  {\normalfont\normalsize\bfseries}}
\makeatother

\definecolor{cvprblue}{rgb}{0.21,0.49,0.74}
\usepackage[breaklinks,colorlinks,allcolors=cvprblue]{hyperref}

\title{ReCoRain: Reliability-Coupled Content and Noise Conditioning for Event-Guided Video Deraining}

\author{
  \begin{tabular}{ccc}
    Runzhe Li & Zhaocheng Yu & Binbin Song \\
    \small Harbin Institute of Technology & \small Harbin Institute of Technology & \small Harbin Institute of Technology \\
    \small \texttt{lirunzhe@stu.hit.edu.cn} & \small \texttt{yuzhaocheng@stu.hit.edu.cn} & \small \texttt{binbinsong@hit.edu.cn} \\[5pt]
    Kui Jiang & Junjun Jiang & Xianming Liu \\
    \small Harbin Institute of Technology & \small Harbin Institute of Technology & \small Harbin Institute of Technology \\
    \small \texttt{jiangkui@hit.edu.cn} & \small \texttt{jiangjunjun@hit.edu.cn} & \small \texttt{csxm@hit.edu.cn}
  \end{tabular}
}

\begin{document}
\maketitle
\begin{abstract}
Event-guided video deraining depends on reliable cross-frame correspondences, yet events generated by scene motion, rain, and sensor noise can provide ambiguous matching evidence.
In diffusion-based restoration, an incorrect match can transfer unrelated content and correlate initial noise between unrelated scene locations.
We propose \emph{ReCoRain}, an event-guided residual video diffusion framework that couples temporal content transfer and noise correlation through a common assessment of correspondence reliability.
This assessment combines RGB appearance, interval-event evidence, and forward--backward motion consistency.
The content path aggregates neighboring features according to alignment reliability, while the noise path uses composed backward motion fields and their associated reliability estimates to regulate initial-noise sharing across the latent sequence.
We use the same conditional noise construction during residual flow-matching training and inference, incorporating temporal noise coordination into the learned restoration process.
Temporal supervision matches motion-compensated inter-frame differences in restored and clean videos under the same teacher-estimated flow and visibility mask, accommodating legitimate appearance changes.
ReCoRain achieves 32.09\,dB PSNR on NTURain-Syn and 28.73\,dB on RainSynLight25.
Enabling reliability gates in both paths reduces warping error by 33.0\% relative to disabling both gates.

\end{abstract}
\section{Introduction}
\label{sec:intro}

Video deraining aims to recover scene details while maintaining temporal coherence under camera and object motion.
Existing methods exploit cross-frame redundancy through alignment, recurrent reconstruction, and spatio-temporal modeling~\cite{chen2018spac,yang2019frame,sun2025vdmamba}.
Their effectiveness depends on correspondences that associate observations of the same scene structure.
Rain obscures appearance and introduces transient changes, making these associations uncertain~\cite{chen2018spac,zhang2025egvd}.
Reliable temporal restoration therefore requires assessing which neighboring observations can support the reconstruction of a given scene structure.

\begin{figure}[t]
\centering
\includegraphics[width=\columnwidth]{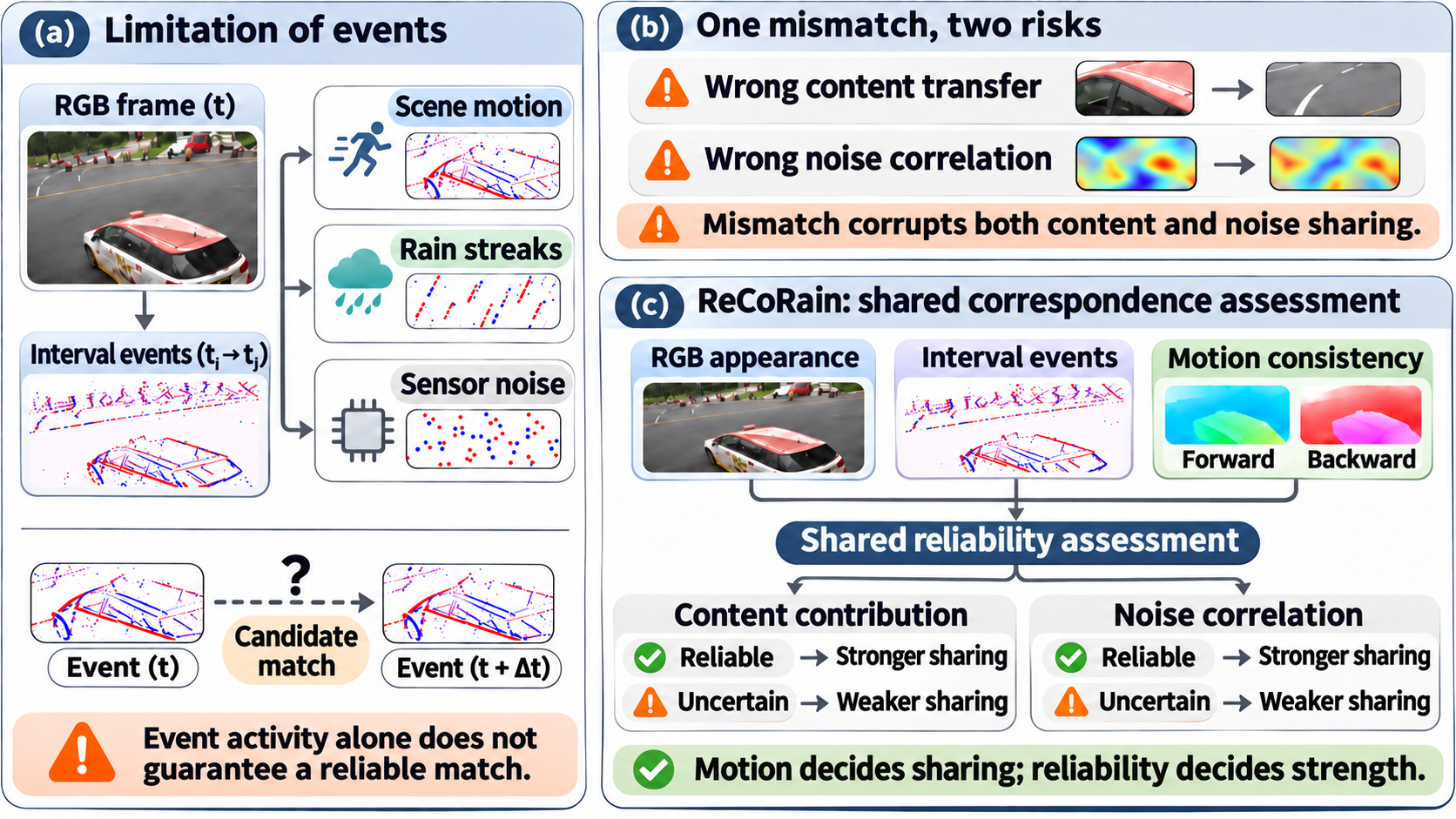}\vspace{-2mm}
% \caption{Conceptual motivation. (a) A plausible match is not a reliable match: rain, scene motion, and sensor noise all produce event activity, so a predicted displacement only proposes a correspondence. (b) One mismatch has two propagation risks: unreliable cross-frame association misdirects both the transferred content and the correlated randomness. (c) One shared reliability assessment supports two selective uses, strengthening both content sharing and noise correlation at supported matches and weakening them at uncertain ones.}
% \label{fig:motivation}
\caption{\textbf{Reliability-coupled content and noise conditioning.}
    (a) Events can arise from scene motion, rain, and sensor noise, so event activity alone does not establish correspondence reliability.
    (b) An incorrect match can transfer unrelated content and correlate initial noise between unrelated scene locations.
    (c) ReCoRain uses a common correspondence assessment to regulate both paths: motion defines candidate associations, and reliability controls the strength of content transfer and noise sharing.}
\vspace{-3mm}
\label{fig:motivation}
\end{figure}

Event cameras record asynchronous brightness changes between RGB frames, providing temporal cues for estimating cross-frame correspondences~\cite{gallego2022survey}.
Event-guided deraining methods exploit these cues for motion-aware aggregation and rain--background discrimination~\cite{sun2023mpevnet,wang2023uvdevent,zhang2025egvd}.
However, events can arise from scene motion, falling rain, and sensor noise~\cite{gallego2022survey,wang2023uvdevent,zhang2025egvd} (Fig.~\ref{fig:motivation}(a)), so event activity alone cannot reliably establish a cross-frame match.
The strength of temporal sharing should therefore depend on evidence supporting the estimated correspondence, rather than on the predicted displacement alone.
Reliability must therefore be assessed from multiple complementary observations, because no single modality remains trustworthy throughout a rainy, dynamic sequence.

Conditional diffusion models combine observation-based guidance with learned priors for detail recovery in image restoration and video weather removal~\cite{liu2024diffplugin,lin2025weatherweaver,varanka2026zsvd}.
In video restoration, temporal consistency depends not only on the scene evidence supplied by conditioning features but also on the temporal correlation of sampled noise~\cite{ge2023pyoco,chang2024warpednoise}.
Independent framewise noise does not explicitly associate corresponding scene locations, whereas sharing noise at fixed image coordinates disregards scene motion.
Motion-based noise transport establishes such associations~\cite{chang2024warpednoise,burgert2025flow}, but their validity depends on the estimated correspondence.
Under rain-induced ambiguity, a mismatch can transfer unrelated content and impose noise correlation between unrelated scene locations (Fig.~\ref{fig:motivation}(b)).
This common failure mode motivates a shared reliability assessment for cross-frame content transfer and noise correlation.

% Diffusion models provide learned priors for detail recovery, with encouraging results in image restoration and video weather removal~\cite{liu2024diffplugin,lin2025weatherweaver,varanka2026zsvd}.
% However, aligning conditioning features and coordinating the random variables used for generation address different aspects of temporal restoration.
% Conditional features provide scene evidence, while prior video-diffusion studies show that the temporal structure of noise also influences the consistency of generated details~\cite{ge2023pyoco,chang2024warpednoise}.
% Independent noise does not explicitly associate corresponding scene locations, whereas noise shared at fixed image coordinates ignores scene motion.
% Motion-based noise transport provides this association~\cite{chang2024warpednoise,burgert2025flow}, but its use in rainy videos must account for unreliable motion estimates. Observation-aware noise modulation has also been explored in other restoration settings: EMR-Diff uses observed edges to modulate noise amplitude for hyperspectral reconstruction~\cite{zhang2026emrdiff}.
% As illustrated in Figure~\ref{fig:motivation}(b), a false correspondence can misdirect both neighboring features and transported noise, associating unrelated scene structures through either process.
% This raises a shared design question: \emph{How should correspondence reliability jointly govern temporal content transfer and cross-frame noise correlation?}

We propose \emph{ReCoRain}, an event-guided residual video diffusion framework with reliability-coupled content and noise conditioning.
Our RGB--event correspondence assessment module (RECA) estimates bidirectional motion and evaluates correspondence reliability using RGB appearance, events recorded over the corresponding frame interval, and forward--backward motion consistency.
Both conditioning paths use this common correspondence estimator and reliability criterion, strengthening sharing at supported matches and attenuating it at uncertain ones (Fig.~\ref{fig:motivation}(c)).

The \emph{content path} aggregates neighboring RGB--event features according to their alignment reliability and injects the resulting evidence into multiple denoising blocks.
The \emph{noise path} uses composed backward motion fields to correlate initial noise across the latent sequence, with the associated reliability estimates controlling the shared component.
We apply the same conditional noise construction during residual flow-matching training~\cite{lipman2023flowmatching} and inference, allowing the model to learn restoration with the temporal noise correlations used at test time.

We supervise temporal behavior by matching motion-compensated inter-frame differences in the restored video to those in the clean reference, using the same flow estimated by a frozen teacher and the same visibility mask.
The reference differences need not be zero, so this objective accommodates legitimate appearance changes rather than forcing aligned frames to be identical.
A bounded wavelet-domain refinement head provides high-frequency corrections after decoding. ReCoRain achieves 32.09\,dB PSNR on NTURain-Syn and 28.73\,dB on RainSynLight25.
Enabling reliability gates in both conditioning paths reduces warping error by 33.0\% relative to disabling both gates.

Our contributions are summarized as follows:
\begin{itemize}
    \item We propose ReCoRain, an event-guided residual video diffusion framework that couples temporal content transfer and initial-noise correlation through a common RGB--event assessment of correspondence reliability.
    %We introduce event-supported dual-path conditioning that uses a common bidirectional correspondence estimator and reliability assessment rule to coordinate temporal feature transfer and cross-frame noise correlation for residual video deraining.
    \item We develop a reliability-conditioned initial-noise construction that uses composed backward motion fields and associated reliability estimates to regulate sharing across the latent sequence, and apply it consistently during residual flow-matching training and inference.
    %We combine RGB appearance, interval-event evidence, and bidirectional motion consistency to assess alignment reliability, providing a common basis for selective content aggregation and adaptive noise sharing.
    \item We formulate temporal supervision by matching inter-frame differences in restored and clean videos with teacher flow and visibility masking, accommodating legitimate temporal changes.
    %We supervise restored inter-frame relations against clean-video relations under the same frozen teacher-flow warping and visibility masking, and evaluate the complete residual restoration framework for reconstruction fidelity and temporal coherence.
\end{itemize}

\section{Related Work}
\label{sec:related}

\subsection{Video Deraining}
Video deraining separates rain from scene content by exploiting their different spatial and temporal structures.
Early approaches capture recurring rain patterns with low-rank priors~\cite{chen2013lowrank}.
Learning-based methods recover cross-frame evidence through superpixel alignment and detail compensation~\cite{chen2018spac}, or flow-guided recurrent reconstruction~\cite{yang2019frame}.
Complementary work models rain dynamics to support semi-supervised learning~\cite{yue2021s2vd}, while state-space architectures expand spatiotemporal modeling and adaptive aggregation~\cite{wu2024rainmamba,sun2025vdmamba}.
Our focus is how correspondence reliability should regulate the use of neighboring observations under rain-induced ambiguity.

\subsection{Event-Guided Restoration}
Event-guided restoration uses asynchronous brightness changes to complement RGB appearance~\cite{gallego2022survey}.
For deraining, research has explored progressive reconstruction, cross-modal fusion, and complementary dense--sparse representations~\cite{sun2023mpevnet,wang2023uvdevent,fu2024ehn}.
EGVD makes temporal aggregation selective through event-aware masks and uses multimodal priors to distinguish rain from background~\cite{zhang2025egvd}.
Diffusion-based extensions address low-light video enhancement and image dehazing~\cite{chen2025evdiffuser,wang2026evdehaze}, as well as long-horizon event-based video reconstruction, prediction, and interpolation~\cite{fan2026longe2v}.
Event activity can nevertheless reflect both scene motion and rain~\cite{wang2023uvdevent,zhang2025egvd}.
ReCoRain therefore treats interval events as one source of correspondence evidence, evaluated jointly with RGB appearance and bidirectional motion consistency.

\begin{figure*}[t]
\centering
\includegraphics[width=0.92\linewidth]{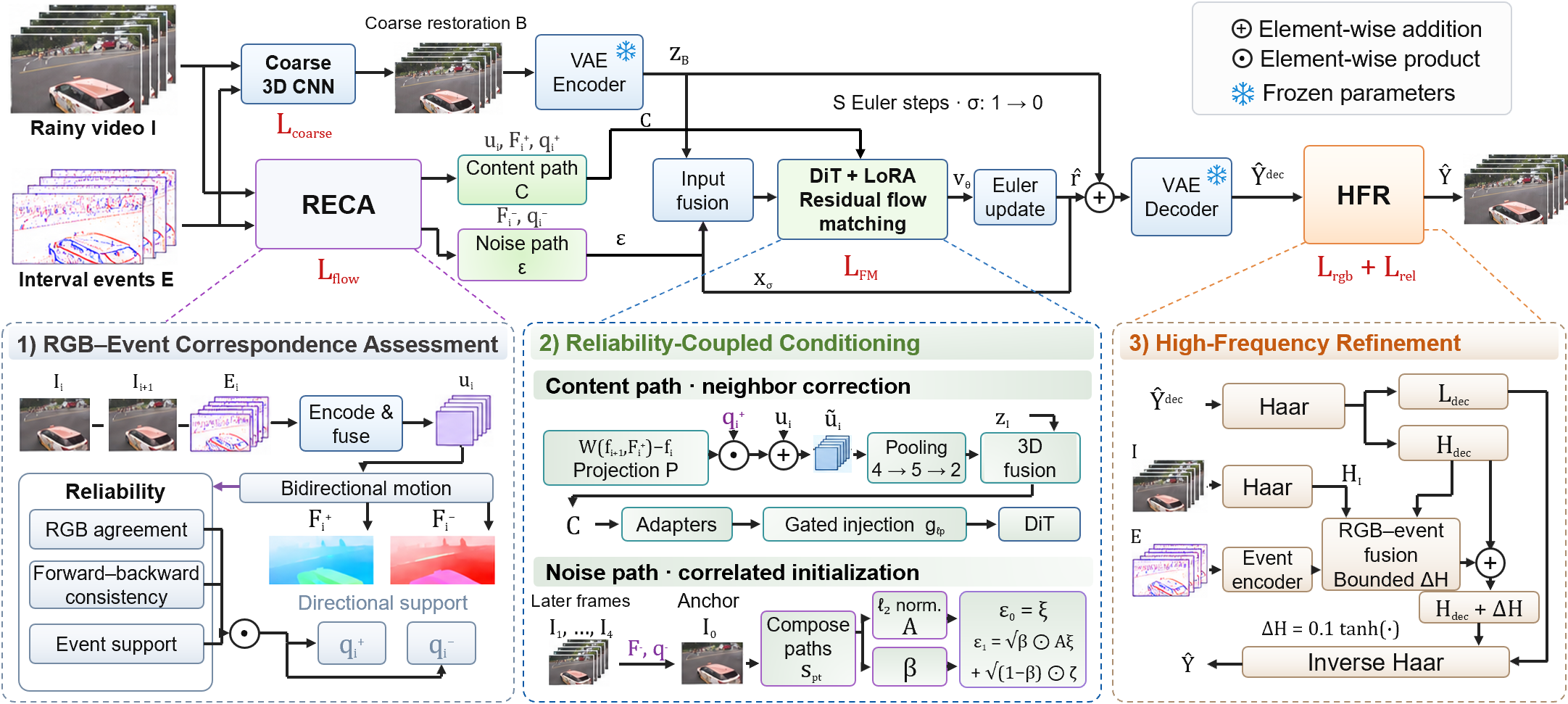}\vspace{-1mm}
\caption{Overview of ReCoRain. RECA provides bidirectional motion and directional reliability estimates, which regulate content transfer and initial-noise correlation. Residual flow matching refines the coarse estimate in latent space, followed by bounded high-frequency refinement after decoding. Red terms indicate training supervision.}\vspace{-3mm}
\label{fig:framework}
\end{figure*}

\subsection{Diffusion Restoration and Noise Modeling}
Diffusion-based restoration combines learned generative priors with conditions that specify degradation or preserve observed content.
For adverse weather, representative approaches use patchwise diffusion~\cite{ozdenizci2023weather}, task-specific and spatial-detail conditioning~\cite{liu2024diffplugin}, and distinct weather and scene prompts~\cite{chen2024t3weather}.
Video methods extend diffusion priors to controllable weather synthesis and removal~\cite{lin2025weatherweaver} and zero-shot deraining through inversion, negative prompting, and attention control~\cite{varanka2026zsvd}.

Beyond conditioning features, diffusion formulations structure the residual and stochastic components of restoration~\cite{liu2024rddm}.
For video generation, shared and frame-specific noise components and correlated noise priors encode temporal dependence~\cite{luo2023videofusion,ge2023pyoco}.
Distribution-preserving noise transport relates stochastic variation across moving image coordinates~\cite{chang2024warpednoise}, while Go-with-the-Flow uses warped noise for motion control and adjusts the strength of motion conditioning through noise degradation~\cite{burgert2025flow}.
% Warped Diffusion formulates video inverse problems in function space and introduces equivariance guidance~\cite{daras2024warpeddiffusion}.
Observation-dependent noise amplitude has been explored in hyperspectral super-resolution using multispectral edges~\cite{zhang2026emrdiff}.
ReCoRain couples content conditioning and noise construction through a common RGB--event assessment of correspondence reliability.
Motion specifies candidate cross-frame associations, while their support controls both neighboring-content transfer and initial-noise correlation.

\section{Method}
\label{sec:method}
Figure~\ref{fig:framework} outlines ReCoRain, which learns a latent correction to a coarse restoration.
A shared RGB--event assessment produces direction-specific reliability weights for neighboring-content correction and motion-based initial-noise correlation.Additional method, implementation, and derivation details are provided in the supplement.

\subsection{Residual Restoration Formulation}
Given five rainy frames $I=\{I_i\}_{i=0}^{4}$ and interval events $E=\{E_i\}_{i=0}^{3}$, we estimate the clean video $Y=\{Y_i\}_{i=0}^{4}$.
An RGB--event network first predicts a coarse restoration $B$.
Using the frozen VAE encoder $\mathcal E$ and decoder $\mathcal D$ of Wan~\cite{wan2025wan}, we define
%\begin{equation}
    $z_B=\mathcal E(B),\qquad
    r=\mathcal E(Y)-z_B.$
    %\label{eq:residual}
%\end{equation}
The five-frame video is encoded into two temporal latent positions.
The residual predictor estimates $\hat r$ under content conditioning and a correspondence-dependent noise initialization.
Decoding gives $\hat Y^{\rm dec}=\mathcal D(z_B+\hat r)$,
which is refined by a high-frequency refinement head (HFR)
to obtain $\hat Y$.
The coarse network is supervised directly in image space; its encoded latent $z_B$ and the target $r$ are detached when training the residual predictor.

% Given five rainy frames $I=\{I_i\}_{i=0}^{4}$ and event measurements $E=\{E_i\}_{i=0}^{3}$ recorded between consecutive frames, we estimate the clean video $Y=\{Y_i\}_{i=0}^{4}$.
% A three-level 3D CNN first predicts a coarse restoration $B$.
% Its RGB and event branches interact through gated residual blocks at each scale.
% A zero-initialized output layer predicts a correction to $I$, so the coarse network starts from the identity mapping and is jointly optimized with the restoration modules.

% We use the frozen VAE encoder $\mathcal E$ and decoder $\mathcal D$ of the pretrained Wan video model~\cite{wan2025wan}.
% The remaining restoration target is defined in latent space:
% \begin{equation}
%  z_B=\mathcal E(B),\qquad r=\mathcal E(Y)-z_B.
%  \label{eq:residual}
% \end{equation}
% This parameterization expresses restoration as a correction to the coarse estimate.
% The five-frame input is represented by two temporal latent positions.
% Two conditioning paths share a correspondence assessment: the content path supplies aligned neighboring evidence, and the noise path coordinates the stochastic initialization.
% Given an estimated residual $\hat r$, decoding yields $\hat B=\mathcal D(z_B+\hat r)$, which is further refined by the high-frequency refinement head (HFR).

\subsection{RGB--Event Correspondence Assessment}
\label{sec:rtc}
% RECA estimates candidate motion and evaluates the observational support for each correspondence.
% Shallow strided convolutions encode the RGB frames and the ten-bin event tensor of each interval.
% For interval $i$, the endpoint RGB features are represented by their mean and absolute difference, and the event features are incorporated through a sigmoid gate.
% Residual 3D convolutions produce an interval representation $u_i$, from which RECA predicts bidirectional displacements.
% Here $F_i^+$ maps locations in frame $i$ toward frame $i+1$, and $F_i^-$ has the reverse direction.
% Both directions are supervised using frozen-teacher flow estimates from clean frame pairs.
RECA encodes the endpoint RGB frames and the events within each interval into a representation $u_i$.
It predicts forward displacements $F_i^+$ from frame $i$ to frame $i+1$ and backward displacements $F_i^-$ in the reverse direction.
Both directions are supervised by flow estimated from clean frame pairs using a frozen teacher.
At inference, correspondence assessment uses only the rainy RGB frames, interval events, and predicted displacements.

Let $\mathcal W(X,F)(x)=X(x+F(x))$ denote bilinear sampling.
For each direction, RECA combines warped RGB agreement, forward--backward consistency, and event support via
%\begin{equation}
    $q_i^\pm
    =
    m_i^\pm\odot
    a_{{\rm rgb},i}^\pm\odot
    a_{{\rm fb},i}^\pm\odot
    a_{{\rm evt},i}^\pm.$
    %\label{eq:reliability}
%\end{equation}
Here $m_i^\pm$ masks out-of-image matches,
$a_{{\rm rgb},i}^\pm=\exp(-d_{{\rm rgb},i}^\pm/\tau_{\rm rgb})$,
and $a_{{\rm fb},i}^\pm=\exp(-e_{{\rm cyc},i}^\pm/\tau_{\rm cyc})$.
The quantities $d_{\rm rgb}$ and $e_{\rm cyc}$ measure local RGB disagreement and the squared motion-cycle residual, respectively.
The event factor is
$a_{{\rm evt},i}^\pm=1-\frac12p_i^\pm(1-s_{{\rm evt},i}^\pm)$,
where $p_i^\pm$ indicates event activity and $s_{{\rm evt},i}^\pm\in[0,1]$ measures time-ordered event support along the proposed trajectory.
This factor lies in $[1/2,1]$ and equals one when no event is present.

The resulting $q_i^\pm$ are bounded support weights rather than calibrated match probabilities.
The common assessment rule produces directional scores: $q_i^+$ controls the explicit neighbor correction, and $q_i^-$ controls backward noise transport.

\subsection{Reliability-Coupled Conditioning}
\label{sec:noise}

\paragraph{Content Correction.}
Let $f_i$ denote the RGB-stem features on the grid of $u_i$.
We form an aligned-neighbor correction on the frame-$i$ reference grid:
\begin{equation}
    \tilde u_i
    =
    u_i+
    q_i^+\odot
    P\!\left(\mathcal W(f_{i+1},F_i^+)-f_i\right),
    \label{eq:content}
\end{equation}
where $P$ is a pointwise projection block.
The support weight regulates the added correction, while the original interval representation remains available.
A fixed temporal pooling maps the four corrected interval features to two condition positions, matching the latent temporal layout.
This pooling performs no additional spatial alignment between interval grids.
The pooled features are fused with $z_I=\mathcal E(I)$ to form a content condition $C$.
Adapters inject $C$ into selected DiT blocks through gates that depend on the noise level and the current token--condition agreement. These gates regulate condition injection; $q_i^+$ regulates the correspondence-based correction used to construct $C$.

\begin{figure}[t]
\centering
\includegraphics[width=\columnwidth]{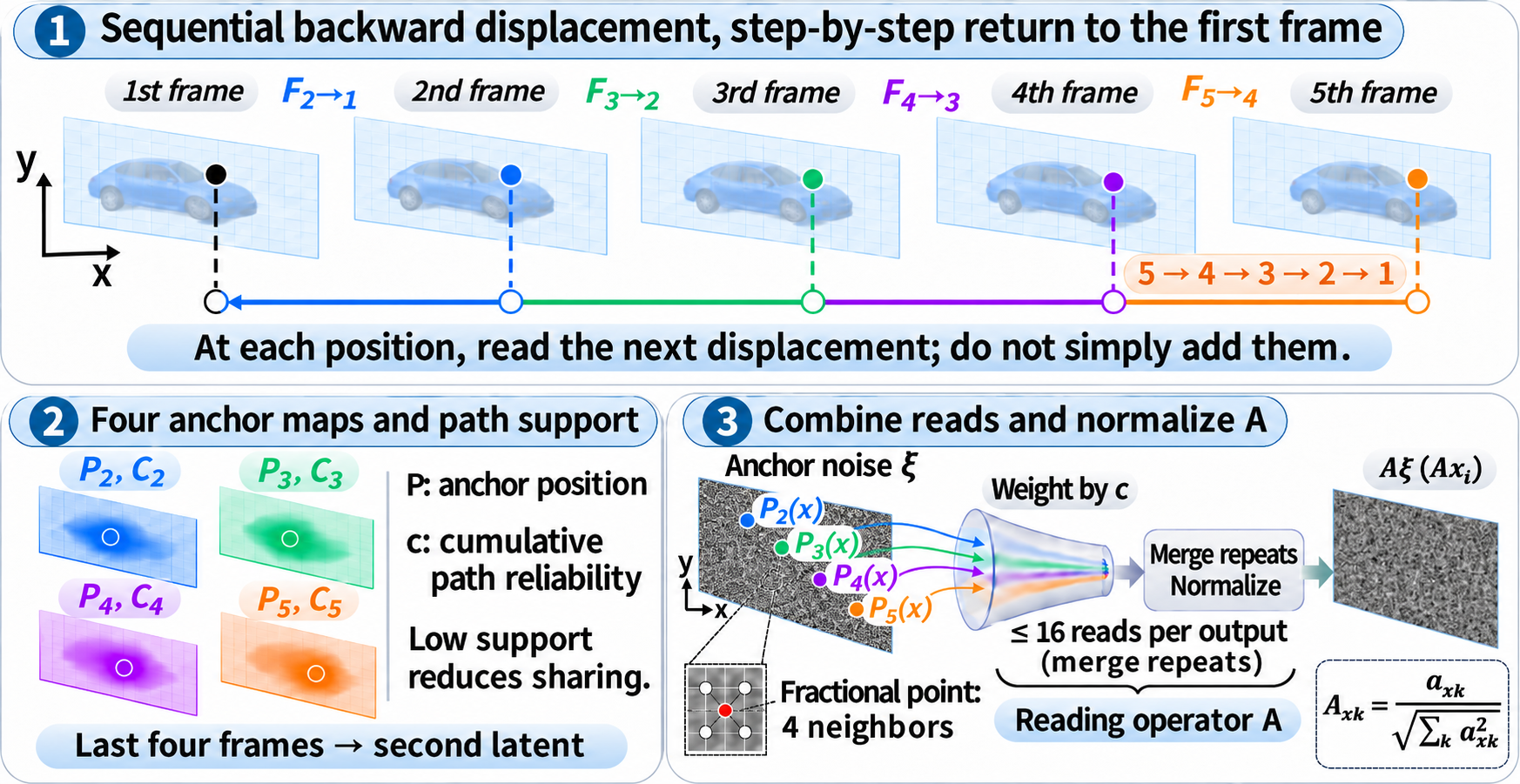}\vspace{-2mm}
% \caption{Noise construction on the latent grid.
% Four backward paths connect frames $1$--$4$ to the frame-$0$ anchor.
% Their accumulated support weights the corresponding bilinear reads, whose repeated anchor indices are merged before row normalization.
% Each supported row of $A$ defines a unit-variance anchor-noise read.
% Mixing this read with independent noise according to $\beta$
% preserves standard-normal coordinate marginals;
% unsupported rows use independent noise.
% The anchor group retains $\xi$.}
\caption{Conditional noise construction. Four backward paths define a sampling row, with repeated anchor indices merged before $\ell_2$ normalization. Reliability-controlled mixing with independent noise preserves standard-normal coordinate marginals; unsupported rows use independent noise.}\vspace{-2mm}

\label{fig:noise-path}
\end{figure}

\paragraph{Conditional Noise Construction.}
As illustrated in Figure~\ref{fig:noise-path}, we compose adjacent
backward displacements to connect frames $1$--$4$ to the frame-$0$ anchor.
Along each path, backward support scores are sampled and multiplied.
For latent location $p$ and starting frame $t\in\{1,\ldots,4\}$, let $s_{pt}$ be the resulting anchor coordinate on the latent grid and $c_{pt}\in[0,1]$ the accumulated support.
Invalid paths have zero support.
Pooling these four paths defines a sampling row for the second latent position.

Let $b_k(s)$ denote the bilinear weight of anchor index $k$ at coordinate $s$.
We sum contributions to repeated anchor indices before normalizing:
\begin{equation}
\begin{aligned}
    w_{pk}&=\sum_{t=1}^{4}c_{pt}b_k(s_{pt}),\\
    A_{pk}&=
    \begin{cases}
        w_{pk}/\|w_{p:}\|_2,&\|w_{p:}\|_2>0,\\
        0,&\text{otherwise}.
    \end{cases}
\end{aligned}
\label{eq:transport}
\end{equation}
The sparse operator $A$ allocates the transported component across anchor coordinates.
The shared variance is set separately by
\begin{equation}
    \bar q_p=\frac14\sum_{t=1}^{4}c_{pt},~~
    %\qquad
    \beta_p=\beta_{\max}\bar q_p,~~
    %\qquad
    \beta_{\max}=0.75,
    \label{eq:sharing}
\end{equation}
with $\beta_p=0$ whenever $A_{p:}=0$.
This is a fixed mapping from accumulated support to shared variance.

We draw mutually independent tensors $[\xi,\zeta]$ with independent standard-normal entries, independently of the observations and their derived transport weights.
Each tensor has the shape of one latent temporal position.
The initial noise $\varepsilon$ is formed by temporally concatenating $\varepsilon_0$ and $\varepsilon_1$:
%The initial noise $\varepsilon$ via the temporal concatenation of $\varepsilon_0$ and $\varepsilon_1$ is 
\begin{equation}
%\begin{aligned}
\varepsilon_0=\xi,~~ \varepsilon_1=\sqrt{\beta}\odot A\xi
              +\sqrt{1-\beta}\odot\zeta,
\varepsilon=(\varepsilon_0,\varepsilon_1).
%\end{aligned}
\label{eq:noise}
\end{equation}
$A$ acts on each channel's spatial grid. % and the last line denotes temporal concatenation.
The cap on $\beta$ retains at least $25\%$ independent variance in each coordinate of the second group.

\paragraph{Conditional Noise Statistics.}
For a fixed observation-derived $A$ and $\beta$, Eq.~\ref{eq:noise} is jointly Gaussian with zero mean.
For each latent channel,
\begin{equation}
\begin{aligned}
&\operatorname{Var}(\varepsilon_{1,p}\mid A,\beta)
=\beta_p\|A_{p:}\|_2^2+1-\beta_p=1,\\
&\operatorname{Cov}(\varepsilon_{1,p},\varepsilon_{0,k}\mid A,\beta)
=\sqrt{\beta_p}A_{pk}.
\end{aligned}
\label{eq:noise_variance}
\end{equation}
These identities also hold for unsupported rows because $A_{p:}=0$ and $\beta_p=0$.
Since all coordinate variances are one,
%\begin{equation}
    $\sum_k
    \operatorname{Corr}^2
    (\varepsilon_{1,p},\varepsilon_{0,k}\mid A,\beta)
    =\beta_p.$
    %\label{eq:noise_coupling}
%\end{equation}
Thus, $A$ determines the relative allocation of correlation, while $\beta_p$ controls the shared variance and the aggregate squared correlation with the anchor group.
The construction preserves standard-normal coordinate marginals while changing the joint noise prior.
Overlapping anchor reads can also introduce spatial correlation within the second group.

\subsection{Residual Flow Matching}
\label{sec:flow_matching}

We use the same conditional noise construction during training and inference.
For a sampled scheduler noise level $\sigma$, the residual path and flow-matching objective~\cite{lipman2023flowmatching} are
\begin{equation}
\begin{aligned}
&x_\sigma=(1-\sigma)r+\sigma\varepsilon,\\
&\mathcal L_{\rm FM}
=\mathbb E\bigl\|
v_\theta(x_\sigma,\sigma,C,z_B)-(\varepsilon-r)
\bigr\|_2^2.
\end{aligned}
\label{eq:fm}
\end{equation}
The target $\varepsilon-r$ is the derivative of the sampled linear path, irrespective of correlations between noise coordinates.
The evolving state remains in residual latent space, with $z_B$ supplied as an additional condition.
For decoded training losses, one velocity prediction yields
\begin{equation}
    \hat r_\sigma
    =
    x_\sigma-\sigma
    v_\theta(x_\sigma,\sigma,C,z_B).
    \label{eq:training-residual}
\end{equation}
This expression recovers $r$ for the exact sampled-path velocity.
We decode $z_B+\hat r_\sigma$ and apply HFR to obtain the training
output $\hat Y$, without unrolling the inference sampler.
At inference, Euler integration starts from $x^{(0)}=\varepsilon$ and follows decreasing noise levels to $\sigma=0$; its final state is $\hat r$.
% The schedule and discrete update are specified in the supplementary material.

% Pretrained DiT and VAE parameters are frozen; LoRA and the added restoration modules are optimized.
% Displacement proposals are detached before assessment and feature sampling, and the noise construction treats $q$, $A$, and $\beta$ as constants.
% The frozen decoder retains gradients with respect to its latent input, allowing decoded-output losses to update the residual predictor, conditioning modules, and refinement head.
% The complete gradient routing is summarized in the supplementary material.

\subsection{Reconstruction and Temporal Supervision}
\label{sec:loss}

\paragraph{High-Frequency Refinement.}
%Let $(L_{\rm dec},H_{\rm dec})=\mathcal H(\hat Y^{\rm dec})$ denote 
We get the $L_{\rm dec}$ and $H_{\rm dec}$ via a one-level Haar decomposition.
HFR predicts a correction from the decoded and rainy high bands together with event features:
\begin{equation}
    \hat Y=
    \mathcal H^{-1}
    (L_{\rm dec},H_{\rm dec}+\Delta H),
    \qquad
    \|\Delta H\|_\infty\le\delta,
    \label{eq:hfr}
\end{equation}
where $\delta=0.10$.
% The decoded low band is retained, and the bound applies to the high-band coefficients.
% The predictor and its $\tanh$ parameterization are detailed in the supplementary material.

\paragraph{Reference Temporal Differences.}
A frozen RAFT teacher~\cite{teed2020raft} estimates bidirectional flow from each clean pair $(Y_i,Y_{i+1})$.
Let $F_i^{*,+}$ denote the forward teacher flow and $M_i$ its validity mask, derived from image bounds and forward--backward consistency.
Using this same flow and mask for the restored and reference videos, we define
\begin{equation}
\begin{aligned}
&R_i(X)=X_i-\mathcal W(X_{i+1},F_i^{*,+}),\\
&\mathcal L_{\rm rel}
=\frac14\sum_{i=0}^{3}
\left\langle\rho\!\left(R_i(\hat Y)-R_i(Y)\right)
\right\rangle_{M_i}.
\end{aligned}
\label{eq:relation}
\end{equation}
where $\rho$ is the Charbonnier function and the masked mean averages over valid pixels and color channels.

To interpret this objective, let $e_i=\hat Y_i-Y_i$ be the restoration error.
Bilinear sampling is linear in its image argument for a fixed flow, giving
\begin{equation}
    R_i(\hat Y)-R_i(Y)
    =
    e_i-\mathcal W(e_{i+1},F_i^{*,+}).
    \label{eq:relation_error}
\end{equation}
The loss therefore penalizes variation in restoration error along teacher-estimated correspondences.
Its target retains the clean video's potentially nonzero temporal differences, accommodating reference appearance changes.
The teacher flow and mask remain estimates; the RGB loss separately constrains reconstruction fidelity.

\paragraph{Training Objective.}
The complete loss is
\begin{equation}
    \mathcal L
    =
    \mathcal L_{\rm FM}
    +\lambda_c\mathcal L_{\rm coarse}
    +\lambda_f\mathcal L_{\rm flow}
    +\lambda_t\mathcal L_{\rm rel}
    +\lambda_r\mathcal L_{\rm rgb}.
    \label{eq:total}
\end{equation}
Here $\mathcal L_{\rm coarse}$ is the L1 loss between $B$ and $Y$;
$\mathcal L_{\rm flow}$ regresses both displacement directions to teacher flow with Smooth L1 over the valid image domain;
and $\mathcal L_{\rm rgb}$ is the MSE between the refined output $\hat Y$ and $Y$.
The RGB and temporal losses use the sampled-state estimate in Eq.~\ref{eq:training-residual}.

\section{Experiments}
\label{sec:experiments}
\begin{table*}[t]
\centering
\scriptsize
\setlength{\tabcolsep}{1.8pt}
\renewcommand{\arraystretch}{1.03}
\caption{Quantitative comparisons on paired synthetic and unpaired real rainy videos. ReCoRain is trained only on NTURain-Syn, without target-dataset fine-tuning or test-time adaptation. Event inputs are simulated with v2e on all splits. Arrows indicate preferred metric directions; bold and underlined values denote the best and second-best results in each row. Metrics are defined in Section~\ref{sec:experimental_setup}.}
\label{tab:main}
\resizebox{\textwidth}{!}{%
\begin{tabular}{ll*{11}{c}}
\toprule
\multirow{2}{*}{Dataset} & \multirow{2}{*}{Metric} & \multicolumn{7}{c}{RGB-only methods} & \multicolumn{4}{c}{Event-guided methods}\\
\cmidrule(lr){3-9}\cmidrule(lr){10-13}
 & &
\shortstack[c]{Weather\\Diffusion\\\cite{ozdenizci2023weather}} &
\shortstack[c]{Diff-Plugin\\\cite{liu2024diffplugin}} &
\shortstack[c]{DiffIR2VR\\\cite{yeh2024diffir2vr}} &
\shortstack[c]{DiffBIR\\\cite{lin2024diffbir}} &
\shortstack[c]{Turtle\\\cite{ghasemabadi2024turtle}} &
\shortstack[c]{DOVE\\\cite{chen2025dove}} &
\shortstack[c]{ZSVD\\\cite{varanka2026zsvd}} &
\shortstack[c]{EGVD-VFI\\\cite{zhang2025egvdvfi}} &
\shortstack[c]{EvDehaze\\\cite{wang2026evdehaze}} &
\shortstack[c]{LongE2V\\\cite{fan2026longe2v}} &
\shortstack[c]{ReCoRain\\(Ours)}\\
\midrule
\multirow[c]{7}{*}{\normalfont NTURain-Syn} & PSNR$\uparrow$ & 25.48 & 24.11 & 23.73 & 25.96 & 27.03 & 26.25 & 25.97 & 27.54 & \underline{30.48} & 29.76 & \textbf{32.09}\\
 & SSIM$\uparrow$ & 0.811 & 0.716 & 0.685 & 0.780 & 0.913 & 0.853 & 0.803 & 0.915 & 0.921 & \underline{0.927} & \textbf{0.930}\\
 & LPIPS$\downarrow$ & 0.147 & 0.114 & 0.276 & 0.122 & 0.071 & 0.133 & 0.194 & 0.058 & \underline{0.044} & 0.050 & \textbf{0.043}\\
 & NIQE$\downarrow$ & 3.475 & 3.841 & 4.221 & 3.685 & \textbf{3.313} & 4.857 & 3.755 & 3.447 & 3.986 & 3.586 & \underline{3.421}\\
 & MANIQA$\uparrow$ & 0.331 & \underline{0.358} & 0.313 & 0.297 & 0.341 & 0.306 & 0.288 & 0.325 & 0.350 & \textbf{0.372} & 0.349\\
 & LIQE$\uparrow$ & 2.741 & \textbf{3.125} & 2.598 & 2.462 & 2.890 & 2.624 & 2.387 & 2.977 & 3.042 & 2.950 & \underline{3.092}\\
 & WE$\downarrow$ & 0.027 & 0.035 & 0.029 & 0.025 & 0.022 & 0.024 & 0.023 & 0.019 & 0.022 & \underline{0.018} & \textbf{0.016}\\
\midrule
\multirow[c]{7}{*}{\normalfont RainSynLight25} & PSNR$\uparrow$ & 24.34 & 21.92 & 22.77 & 23.74 & 26.10 & 25.57 & 21.79 & 25.84 & \underline{27.69} & 26.43 & \textbf{28.73}\\
 & SSIM$\uparrow$ & 0.839 & 0.637 & 0.631 & 0.707 & 0.838 & 0.773 & 0.639 & 0.822 & \underline{0.854} & 0.843 & \textbf{0.868}\\
 & LPIPS$\downarrow$ & 0.243 & 0.196 & 0.366 & 0.291 & 0.212 & 0.266 & 0.362 & \underline{0.186} & 0.191 & 0.198 & \textbf{0.174}\\
 & NIQE$\downarrow$ & 4.529 & \underline{3.752} & 5.166 & 4.530 & 3.943 & 5.123 & 4.408 & 3.987 & 3.825 & 3.892 & \textbf{3.612}\\
 & MANIQA$\uparrow$ & 0.352 & 0.372 & \textbf{0.406} & 0.329 & 0.383 & 0.312 & 0.297 & 0.361 & 0.392 & 0.399 & \underline{0.404}\\
 & LIQE$\uparrow$ & 3.013 & 2.845 & 2.693 & 2.515 & 2.930 & 2.772 & 2.439 & 3.186 & \underline{3.352} & 3.107 & \textbf{3.482}\\
 & WE$\downarrow$ & 0.053 & 0.044 & 0.050 & 0.047 & 0.030 & 0.026 & 0.045 & 0.022 & 0.027 & \underline{0.021} & \textbf{0.019}\\
\midrule
\multirow[c]{7}{*}{\normalfont RainSynComplex25} & PSNR$\uparrow$ & 16.04 & 16.74 & 15.29 & 17.04 & 18.06 & 18.54 & 15.31 & 17.78 & 18.50 & \underline{18.63} & \textbf{18.96}\\
 & SSIM$\uparrow$ & 0.527 & 0.465 & 0.417 & 0.519 & 0.563 & \underline{0.581} & 0.386 & 0.552 & 0.575 & 0.558 & \textbf{0.604}\\
 & LPIPS$\downarrow$ & 0.458 & \underline{0.344} & 0.522 & 0.424 & 0.387 & 0.404 & 0.543 & 0.392 & 0.376 & 0.385 & \textbf{0.336}\\
 & NIQE$\downarrow$ & 8.003 & \textbf{3.867} & 8.318 & 5.209 & \underline{4.543} & 5.613 & 6.632 & 5.607 & 5.363 & 5.497 & 4.965\\
 & MANIQA$\uparrow$ & 0.285 & 0.317 & 0.303 & 0.242 & \textbf{0.369} & 0.295 & 0.209 & 0.311 & 0.335 & 0.342 & \underline{0.353}\\
 & LIQE$\uparrow$ & 2.647 & \underline{3.042} & 2.433 & 2.287 & 2.794 & 2.660 & 2.154 & 2.890 & 2.922 & 2.856 & \textbf{3.191}\\
 & WE$\downarrow$ & 0.144 & 0.078 & 0.118 & 0.112 & 0.058 & \underline{0.045} & 0.130 & 0.059 & 0.052 & 0.076 & \textbf{0.037}\\
\midrule
\multirow[c]{4}{*}{\normalfont NTURain-Real} & NIQE$\downarrow$ & 3.378 & 3.720 & 4.269 & 3.934 & \underline{3.242} & 4.929 & 3.915 & 3.248 & 3.333 & 3.398 & \textbf{3.219}\\
 & MANIQA$\uparrow$ & 0.422 & \textbf{0.494} & 0.389 & 0.363 & 0.447 & 0.395 & 0.372 & 0.461 & 0.452 & 0.439 & \underline{0.474}\\
 & LIQE$\uparrow$ & 3.128 & \textbf{3.574} & 2.894 & 2.742 & 3.296 & 2.616 & 2.804 & 3.411 & 3.342 & 3.269 & \underline{3.484}\\
 & WE$\downarrow$ & 0.033 & 0.040 & 0.028 & 0.033 & 0.025 & \underline{0.021} & 0.026 & 0.023 & 0.029 & 0.022 & \textbf{0.019}\\
\bottomrule
\end{tabular}%
}
\end{table*}

\subsection{Experimental Setup}
\label{sec:experimental_setup}
\paragraph{Datasets and Evaluation Protocol.}
We evaluate on NTURain~\cite{chen2018spac}, RainSynLight25~\cite{yang2019frame}, and RainSynComplex25~\cite{liu2018joint}.
NTURain contains paired synthetic sequences (NTURain-Syn) and unpaired real rainy videos (NTURain-Real); both RainSyn datasets provide paired clean references.
ReCoRain is trained only on NTURain-Syn and evaluated on all test splits without target-dataset fine-tuning or test-time adaptation.
Event streams are simulated from the RGB sequences using v2e~\cite{hu2021v2e} and encoded as temporally binned voxels.
This protocol uses simulated events on every split, including NTURain-Real.

\paragraph{Evaluation Metrics.}
On paired data, we report PSNR~\cite{huynh2008scope}, SSIM~\cite{wang2004ssim}, and LPIPS~\cite{zhang2018lpips} to assess reconstruction quality against clean references.
On all splits, we report NIQE~\cite{mittal2013niqe}, MANIQA~\cite{yang2022maniqa}, and LIQE~\cite{zhang2023liqe} for no-reference image quality, together with motion-compensated temporal warping error (WE)~\cite{lai2018blind}.
We assess temporal consistency jointly with reconstruction and perceptual quality.

\paragraph{Compared Methods.}
We compare with ten restoration baselines.
RGB-only methods include WeatherDiffusion~\cite{ozdenizci2023weather}, Diff-Plugin~\cite{liu2024diffplugin}, DiffIR2VR~\cite{yeh2024diffir2vr}, DiffBIR~\cite{lin2024diffbir}, Turtle~\cite{ghasemabadi2024turtle}, DOVE~\cite{chen2025dove}, and ZSVD~\cite{varanka2026zsvd}.
Event-guided methods include EGVD-VFI~\cite{zhang2025egvdvfi}, EvDehaze~\cite{wang2026evdehaze}, and LongE2V~\cite{fan2026longe2v}.
All methods are evaluated on the same test splits at full spatial resolution.

\paragraph{Implementation Details.}
We adapt Wan2.1-T2V-1.3B~\cite{wan2025wan} by optimizing LoRA~\cite{hu2022lora} parameters and the added restoration modules, while freezing the pretrained backbone weights and VAE.
Training uses five-frame clips with four event intervals and $128\times128$ crops for 50 epochs on one NVIDIA RTX 3090.
We use AdamW with a learning rate of $10^{-4}$, weight decay of $10^{-4}$, and batch size one.
The reliability-conditioned initial-noise construction is used during both residual flow-matching training and inference.
Inference uses 50 Euler steps without classifier-free guidance.
Further implementation details are provided in the supplementary material.
% \paragraph{Implementation.}
% We adapt Wan2.1-T2V-1.3B~\cite{wan2025wan} using LoRA~\cite{hu2022lora} and the added restoration modules, keeping the pretrained backbone and VAE frozen.
% Training uses five-frame clips with four event intervals and $128\times128$ crops for 50 epochs on one NVIDIA RTX 3090.
% We use AdamW with learning rate $10^{-4}$, weight decay $10^{-4}$, and batch size one.
% Inference uses 50 Euler steps without classifier-free guidance; further settings are provided in the supplementary material.

\subsection{Comparison with State-of-the-Art Methods}
\label{sec:comparison}
\label{sec:quantitative}
\label{sec:qualitative}

\paragraph{Paired Synthetic Benchmarks.}
As shown in Table~\ref{tab:main}, ReCoRain achieves the best PSNR, SSIM, and LPIPS and the lowest WE among the compared methods on all three paired datasets.
It reaches 32.09, 28.73, and 18.96\,dB PSNR on NTURain-Syn, RainSynLight25, and RainSynComplex25, respectively, exceeding the strongest PSNR baseline on each dataset by 1.61, 1.04, and 0.33\,dB.
The gains on the two unseen RainSyn datasets support cross-dataset transfer of the NTURain-Syn-trained model under the stated protocol.
No-reference metrics give less uniform rankings: ReCoRain leads NIQE and LIQE on RainSynLight25 and LIQE on RainSynComplex25, while other methods lead the remaining no-reference metrics.

\begin{figure*}[t]
  \centering
  \includegraphics[width=\linewidth]{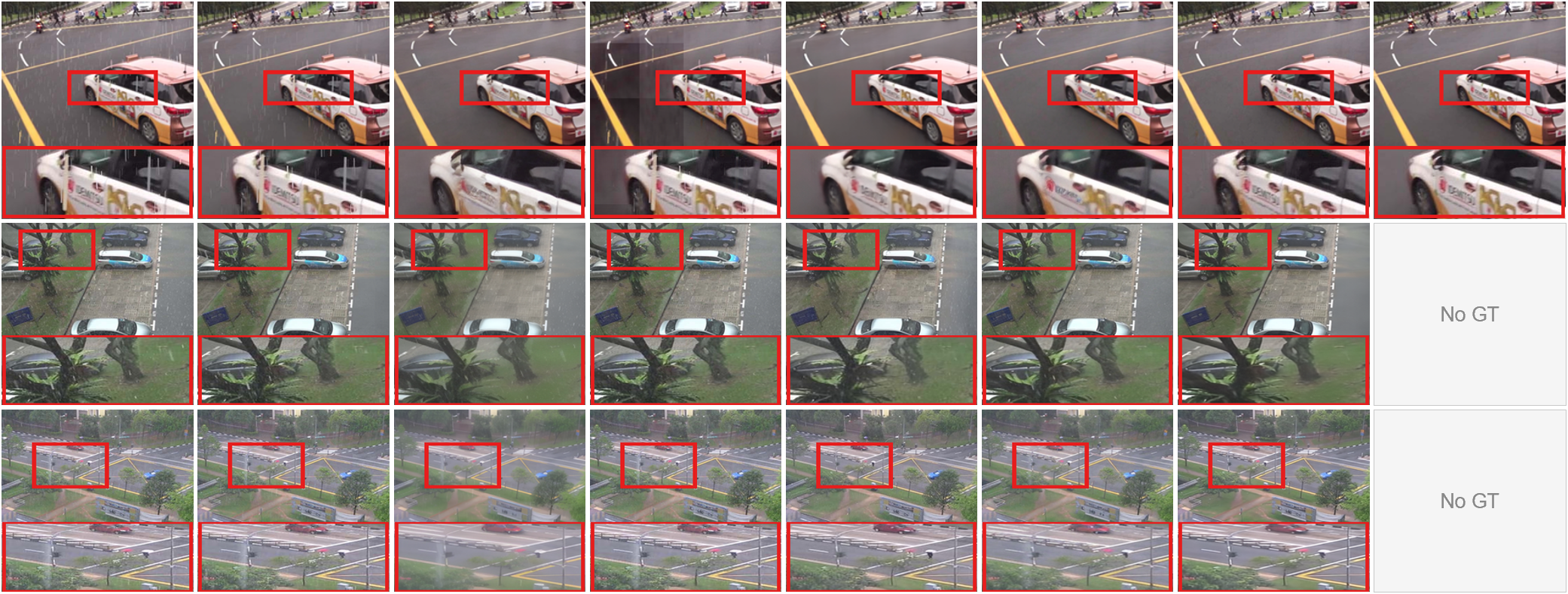}\vspace{-2mm}
  \vspace{1pt}
  {\scriptsize
  \setlength{\tabcolsep}{0pt}%
  \begin{tabular}{@{}*{8}{c}@{}}
    \makebox[0.125\linewidth]{Rainy} &
    \makebox[0.125\linewidth]{WeatherDiffusion~\cite{ozdenizci2023weather}} &
    \makebox[0.125\linewidth]{Diff-Plugin~\cite{liu2024diffplugin}} &
    \makebox[0.125\linewidth]{Turtle~\cite{ghasemabadi2024turtle}} &
    \makebox[0.125\linewidth]{DOVE~\cite{chen2025dove}} &
    \makebox[0.125\linewidth]{EvDehaze~\cite{wang2026evdehaze}} &
    \makebox[0.125\linewidth]{ReCoRain (Ours)} &
    \makebox[0.125\linewidth]{GT} \\
  \end{tabular}}
  \caption{Comparisons on NTURain-Syn (top) and NTURain-Real (bottom two rows). Red-box regions are enlarged below each image. GT is unavailable for NTURain-Real.}\vspace{-2mm}
  \label{fig:qualitative_nturain}
\end{figure*}

Figure~\ref{fig:qualitative_nturain} (top) illustrates the reconstruction differences on NTURain-Syn.
In the selected regions, WeatherDiffusion retains rain streaks, Turtle exhibits windowing artifacts, and Diff-Plugin, DOVE, and EvDehaze blur vehicle markings and lettering.
ReCoRain suppresses the visible streaks while recovering these structures more closely to the clean reference.
% \paragraph{Paired Synthetic Data.}
% ReCoRain achieves the best PSNR, SSIM, LPIPS, and WE on all three paired datasets (Table~\ref{tab:main}).
% Its PSNR gains over the strongest baseline are 1.61, 1.04, and 0.33\,dB on NTURain-Syn, RainSynLight25, and RainSynComplex25, respectively.
% The gains on the two unseen RainSyn datasets support cross-dataset generalization under the fixed training protocol.
% Blind metrics yield different rankings: ReCoRain leads NIQE and LIQE on RainSynLight25 and LIQE on RainSynComplex25, but does not dominate every blind metric.
% In Figure~\ref{fig:qualitative_nturain} (top), WeatherDiffusion leaves rain streaks, Turtle exhibits windowing artifacts, and Diff-Plugin, DOVE, and EvDehaze blur vehicle graphics and lettering.
% ReCoRain suppresses rain while preserving these details closer to GT, consistent with its reference-based fidelity scores.

\paragraph{Unpaired Real Rainy Videos.}
On NTURain-Real, ReCoRain achieves the lowest NIQE and WE and ranks second on MANIQA and LIQE, for which Diff-Plugin ranks first (Table~\ref{tab:main}).
In the selected regions of Figure~\ref{fig:qualitative_nturain} (bottom), WeatherDiffusion, Diff-Plugin, Turtle, and EvDehaze retain visible rain artifacts; Diff-Plugin also blurs scene details.
DOVE removes many streaks but leaves diffuse remnants.
ReCoRain produces fewer visible streaks and sharper scene contours in these examples.

% \paragraph{Unpaired Real-world Data.}
% On NTURain-Real, ReCoRain achieves the lowest NIQE and WE and the second-highest MANIQA and LIQE; Diff-Plugin leads the latter two metrics.
% In Figure~\ref{fig:qualitative_nturain} (bottom), WeatherDiffusion, Diff-Plugin, Turtle, and EvDehaze leave visible rain artifacts, with Diff-Plugin also producing blurrier restorations.
% DOVE reduces rain but retains diffuse remnants.
% ReCoRain removes visible streaks while preserving sharper scene contours in the selected regions.
% The differing blind-metric rankings and visual results emphasize complementary aspects of restoration quality.

\subsection{Ablation Studies}
\label{sec:ablation}
All ablations are evaluated on NTURain-Syn using the same test split, spatial resolution, and metrics as the main comparison.
We examine the correspondence assessment, its use in the two conditioning paths, and the contributions of noise construction and temporal supervision.

\begin{figure}[tbp]
\centering
\includegraphics[width=\linewidth]{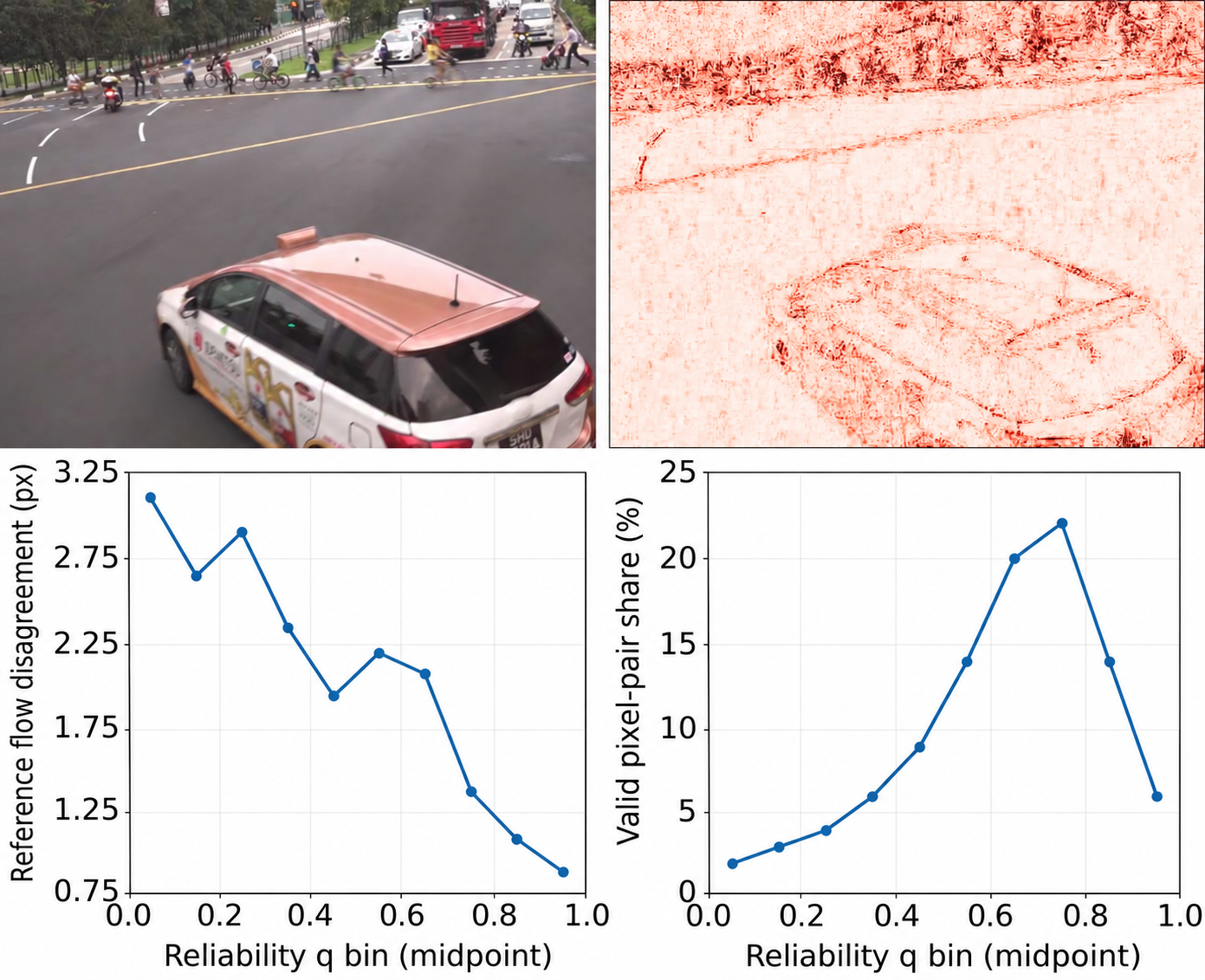}\vspace{-2mm}
\caption{Correspondence-reliability diagnostic. Top: full-model output and its difference from the output without $q$ weighting. Bottom: mean deviation from clean-frame RAFT estimates (pixels) and valid pixel-pair proportions, grouped by $q$. The RAFT estimates serve as a proxy motion reference.}\vspace{-2mm}
%Reliability diagnostic. Top: full-model output and its difference from the output without $q$ weighting. Bottom: mean deviation from clean-frame RAFT flow (pixels) and valid pixel-pair proportions, grouped by $q$.}
\label{fig:ablation_reliability}
\end{figure}

\paragraph{Correspondence Reliability.}
To assess RECA's correspondence support $q$, we group valid pixel pairs by $q$ and measure the mean deviation between their estimated displacements and RAFT flow computed on clean frames (Figure~\ref{fig:ablation_reliability}).
Higher-$q$ bins exhibit smaller deviations from this proxy motion reference; the accompanying proportions show how valid pairs are distributed across the bins.
This association supports using $q$ to regulate the strength of correspondence-based sharing.
The output-difference map shows that $q$ weighting primarily affects moving regions and boundaries in the displayed example.
%Figure~\ref{fig:ablation_reliability} relates correspondence support $q$ to deviation from clean-frame RAFT flow. Higher-support bins have lower deviation, while bin proportions show valid-pair distribution. These diagnostics support using $q$ to regulate correspondence-based sharing. The trend supports interpreting $q$: a high value indicates agreement among appearance, event support, and bidirectional motion evidence, whereas a low value indicates that some evidence questions the proposed match. The output-difference map shows that reliability weighting mainly changes moving regions and boundaries. Thus, $q$ is an observation-supported modulation signal.

\begin{figure}[tbp]
\centering
\includegraphics[width=\linewidth]{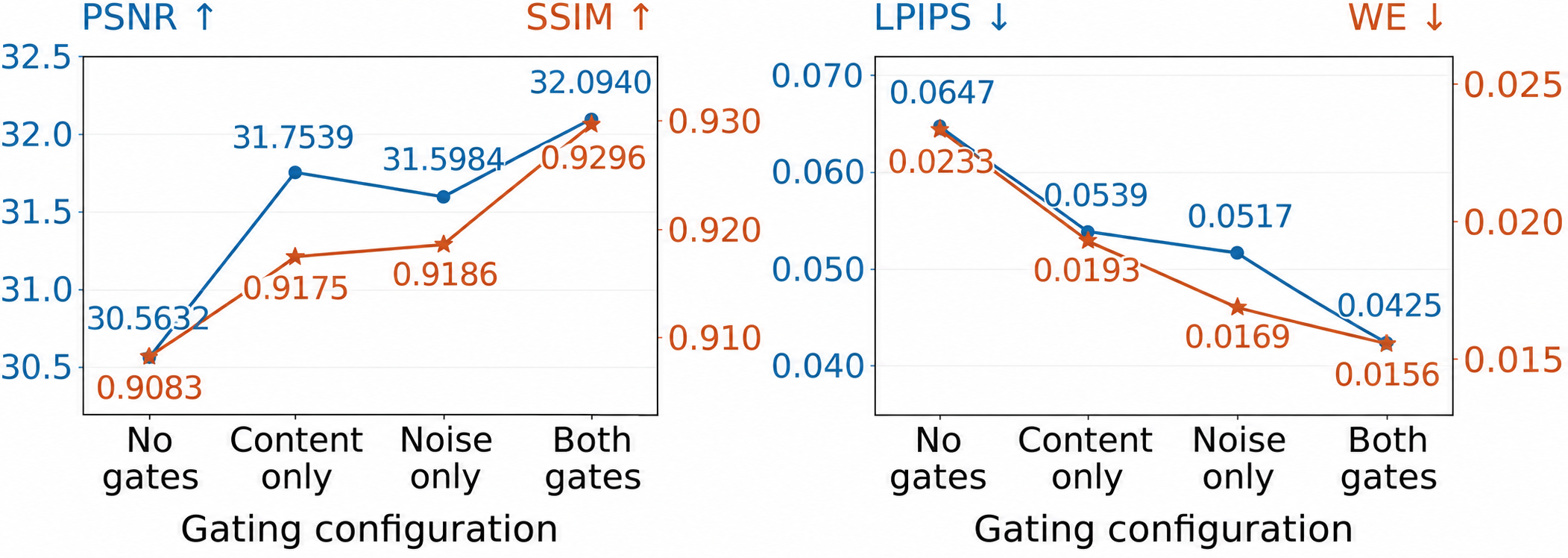}\vspace{-2mm}
\caption{Fixed-motion ablation of $q$-based gating on neither path, the content path, the noise path, or both paths.}\vspace{-2mm}
\label{fig:ablation_gating}
\end{figure}

\paragraph{Reliability Gating in Both Paths.}
With motion estimates fixed, we apply $q$-based gating to neither path, the content path only, the noise path only, or both paths (Figure~\ref{fig:ablation_gating}).
Content-only gating gives higher PSNR than noise-only gating (31.75 versus 31.60\,dB), while noise-only gating yields lower WE (0.0169 versus 0.0193).
Applying the reliability assessment to both paths gives the best results across all four metrics, increasing PSNR by 1.53\,dB and reducing WE by 33.0\% relative to disabling both gates.
These results support the complementary effects of reliability-controlled content transfer and initial-noise correlation.

\paragraph{Reliability-conditioned Initial Noise.}
With the content condition fixed, Table~\ref{tab:ablation_noise} compares independent noise, anchor noise shared at fixed coordinates, motion-transported anchor noise, and reliability-adjusted transport.
Motion transport follows composed backward correspondences across the latent sequence; reliability adjustment additionally weights the transported reads and their sharing strength.
The motion-transported control uses a motion-feature gate to determine correlation strength; the content-only configuration in Figure~\ref{fig:ablation_gating} instead uses ungated correlated noise.

\begin{figure}[tbp]
\centering
\includegraphics[width=\linewidth]{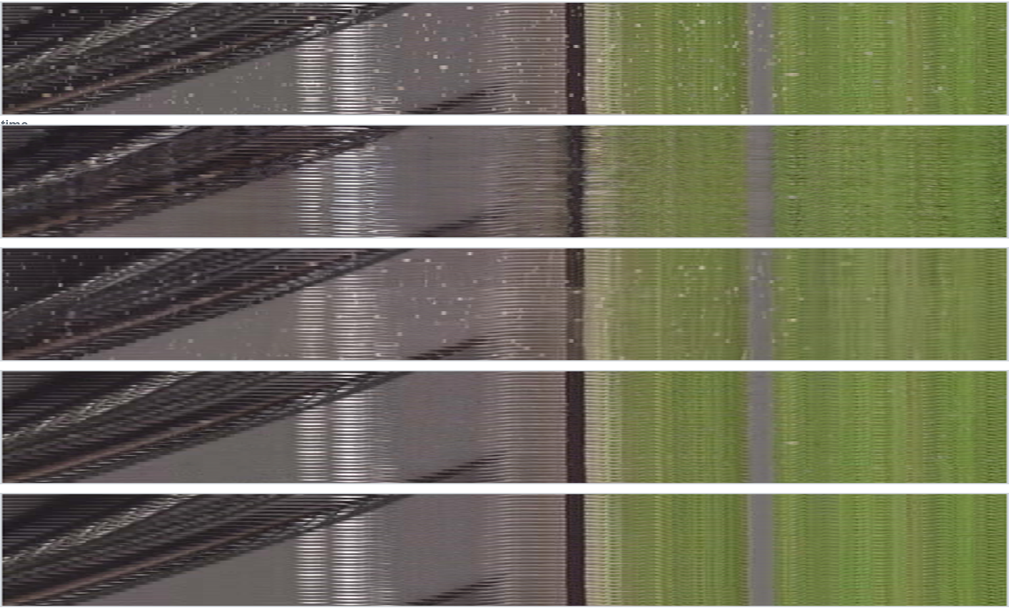}
\caption{Spatiotemporal slices at the same location and frame range. Top to bottom: independent noise, fixed-coordinate sharing, motion transport, reliability-adjusted transport, and GT.}
\label{fig:ablation_noise}
\end{figure}

Motion transport reduces WE from 0.0197 to 0.0172 relative to fixed-coordinate sharing, but slightly worsens SSIM and LPIPS.
Adding reliability adjustment improves all four metrics over motion transport, increasing PSNR by 0.45\,dB and further reducing WE to 0.0156.
The spatiotemporal slices in Figure~\ref{fig:ablation_noise} show trajectories closer to the clean reference after reliability adjustment.
Together, these results support assessing correspondence reliability when using motion to correlate initial noise.

% \paragraph{Noise Construction.}
% With the content condition fixed, we compare independent noise, fixed-coordinate sharing, motion transport, and reliability-adjusted transport.
% Motion transport follows backward correspondences, and reliability adjustment weights both the transported reads and their sharing strength.
% In Table~\ref{tab:ablation_noise}, motion transport reduces WE from 0.0197 to 0.0172 relative to fixed-coordinate sharing.
% Reliability adjustment further improves PSNR by 0.4504\,dB and reduces WE to 0.0156.
% The closer-to-GT trajectories in Figure~\ref{fig:ablation_noise} support the benefit of alignment and correspondence selection beyond correlation alone.

\begin{table}[tbp]
\centering
\footnotesize
\setlength{\tabcolsep}{3pt}
\caption{Initial-noise construction on NTURain-Syn with the content condition fixed. Independent noise removes cross-frame sharing; fixed-coordinate sharing reuses anchor noise without motion transport; motion transport uses backward flow and a motion-feature gate for correlation strength, without $q$ weighting in the noise path. The full model adds reliability weighting to transported reads and sharing strength. }
%Noise-construction ablation on NTURain-Syn with fixed content. ``w/o sharing'' uses independent noise; ``w/o motion'' shares anchor noise at fixed coordinates; ``w/o reliability'' transports anchor noise along backward flow without reliability weighting; ``Full model'' adds reliability adjustment. Best results are in bold.}
\label{tab:ablation_noise}
\begin{tabular}{@{}lcccc@{}}
\toprule
Setting & PSNR$\uparrow$ & SSIM$\uparrow$ & LPIPS$\downarrow$ & WE$\downarrow$\\
\midrule
w/o sharing & 30.8614 & 0.9104 & 0.0573 & 0.0228\\
w/o motion & 31.5728 & 0.9207 & 0.0462 & 0.0197\\
w/o reliability & 31.6436 & 0.9198 & 0.0493 & 0.0172\\
\midrule
Full model & \textbf{32.0940} & \textbf{0.9296} & \textbf{0.0425} & \textbf{0.0156}\\
\bottomrule
\end{tabular}
\end{table}

\begin{figure}[tbp]
\centering
\includegraphics[width=0.92\linewidth]{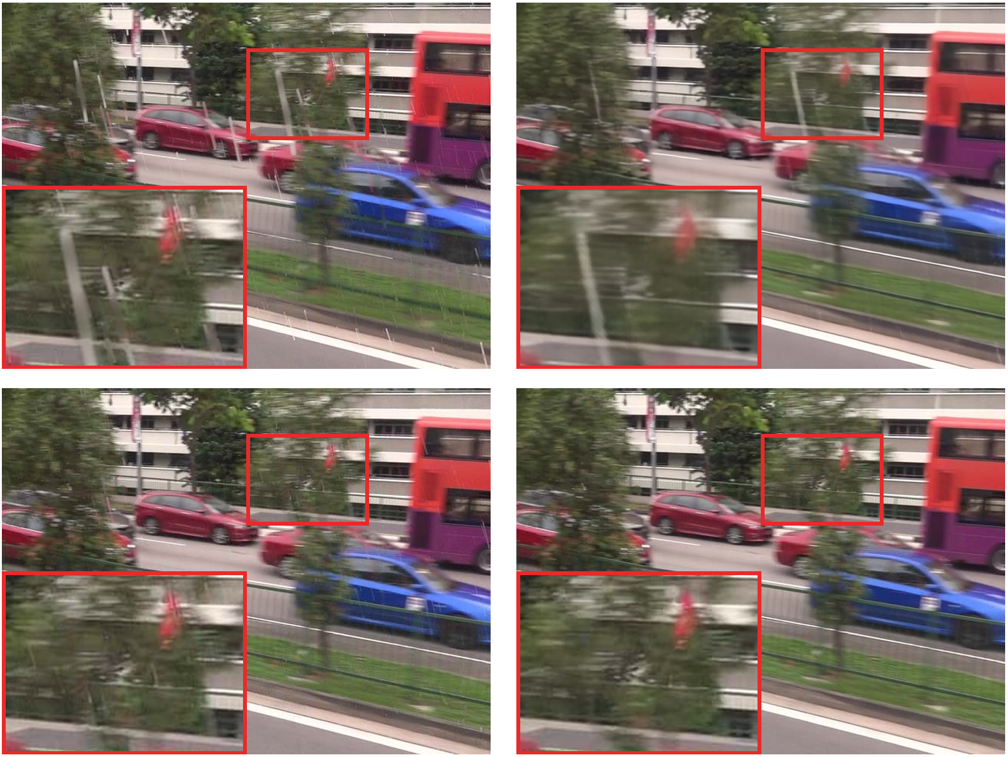}
\caption{Event ablation, in reading order: RGB-only, RGB--event without event evidence in $q$, full RGB--event, and GT. Red boxes mark enlarged regions.}
\label{fig:ablation_event}
\end{figure}

\paragraph{Event Inputs and Reliability Evidence.}
Table~\ref{tab:ablation_event} evaluates the event input and its role in correspondence assessment.
Adding simulated event inputs while excluding event evidence $q_{\mathrm{evt}}$ from $q$ improves PSNR by 2.63\,dB over the RGB-only variant.
Including $q_{\mathrm{evt}}$ yields a further 1.52\,dB gain and reduces WE from 0.0186 to 0.0156.
The latter comparison shows the additional value of event evidence in reliability assessment when event features are already available to the restoration pipeline.
Figure~\ref{fig:ablation_event} shows fewer rain remnants and clearer structures for the full configuration in the displayed regions.

\begin{table}[tbp]
\centering
\footnotesize
\setlength{\tabcolsep}{3pt}
\caption{Event-modality ablation on NTURain-Syn. ``RGB-only'' uses no event input; ``RGB+E (w/o $q_{\mathrm{evt}}$)'' retains event features but excludes event evidence from the reliability score $q$; ``Full model'' uses both event features and event evidence.}
\label{tab:ablation_event}
\begin{tabular}{@{}lcccc@{}}
\toprule
Setting & PSNR$\uparrow$ & SSIM$\uparrow$ & LPIPS$\downarrow$ & WE$\downarrow$\\
\midrule
RGB-only & 27.9473 & 0.9092 & 0.0596 & 0.0241\\
RGB+E (w/o $q_{\mathrm{evt}}$) & 30.5769 & 0.9203 & 0.0542 & 0.0186\\
\midrule
Full model & \textbf{32.0940} & \textbf{0.9296} & \textbf{0.0425} & \textbf{0.0156}\\
\bottomrule
\end{tabular}
\end{table}

% \paragraph{Supervision and Loss Weights.}
% Table~\ref{tab:ablation_loss} removes each auxiliary loss and tests equal weights, with the flow-matching weight fixed at one. The coarse loss stabilizes the residual, the flow loss preserves estimated motion structure, the temporal-relation loss constrains motion-compensated differences, and the RGB loss improves pixel fidelity. The default weights give the best four metrics among settings. Removing temporal-relation supervision increases WE from 0.0156 to 0.0198, showing that temporal supervision contributes beyond per-frame reconstruction. Removing RGB supervision causes the largest PSNR drop (0.6814\,dB). Equal weights also underperform, supporting the selected balance between reconstruction fidelity and temporal consistency.

\paragraph{Temporal Supervision and Loss Weights.}
Table~\ref{tab:ablation_loss} evaluates the removal of each auxiliary loss and an equal-weight setting, with the flow-matching loss weight fixed at one.
The temporal-relation term matches restored and clean inter-frame differences using the same flow from a frozen teacher and the same visibility mask.
Removing this term increases WE from 0.0156 to 0.0198, showing its contribution alongside the remaining reconstruction and motion supervision.
Among the individual loss removals, omitting RGB supervision causes the largest PSNR drop (0.68\,dB).
The default weights give the best results among the tested configurations; equal weighting underperforms in all four metrics.

\begin{table}[!t]
\centering
\scriptsize
\setlength{\tabcolsep}{3.2pt}
\renewcommand{\arraystretch}{1.02}
\caption{Auxiliary-loss ablation on NTURain-Syn. The weights $(\lambda_c,\lambda_f,\lambda_t,\lambda_r)$ correspond to coarse reconstruction, flow supervision, temporal relations, and RGB reconstruction, respectively. The flow-matching loss has unit weight. The last row is the default.}
%Auxiliary-loss weights on NTURain-Syn. The flow-matching weight is one; the last row is the default. Best results are in bold.}
\label{tab:ablation_loss}
\begin{tabular}{@{}ccccc@{}}
\toprule
$(\lambda_c,~~\lambda_f,~~\lambda_t,~~\lambda_r)$ & PSNR$\uparrow$ & SSIM$\uparrow$ & LPIPS$\downarrow$ & WE$\downarrow$\\
\midrule
(0.00,~~0.03,~~0.05,~~0.10) & 31.6421 & 0.9203 & 0.0508 & 0.0187\\
(0.20,~~0.00,~~0.05,~~0.10) & 31.8567 & 0.9234 & 0.0486 & 0.0197\\
(0.20,~~0.03,~~0.00,~~0.10) & 31.9015 & 0.9226 & 0.0499 & 0.0198\\
(0.20,~~0.03,~~0.05,~~0.00) & 31.4126 & 0.9142 & 0.0534 & 0.0192\\
(1.00,~~1.00,~~1.00,~~1.00) & 31.5384 & 0.9157 & 0.0554 & 0.0209\\
\midrule
(0.20,~~0.03,~~0.05,~~0.10) & \textbf{32.0940} & \textbf{0.9296} & \textbf{0.0425} & \textbf{0.0156}\\
\bottomrule
\end{tabular}
\end{table}

\begin{figure}[tbp]
\centering
\includegraphics[width=\linewidth]{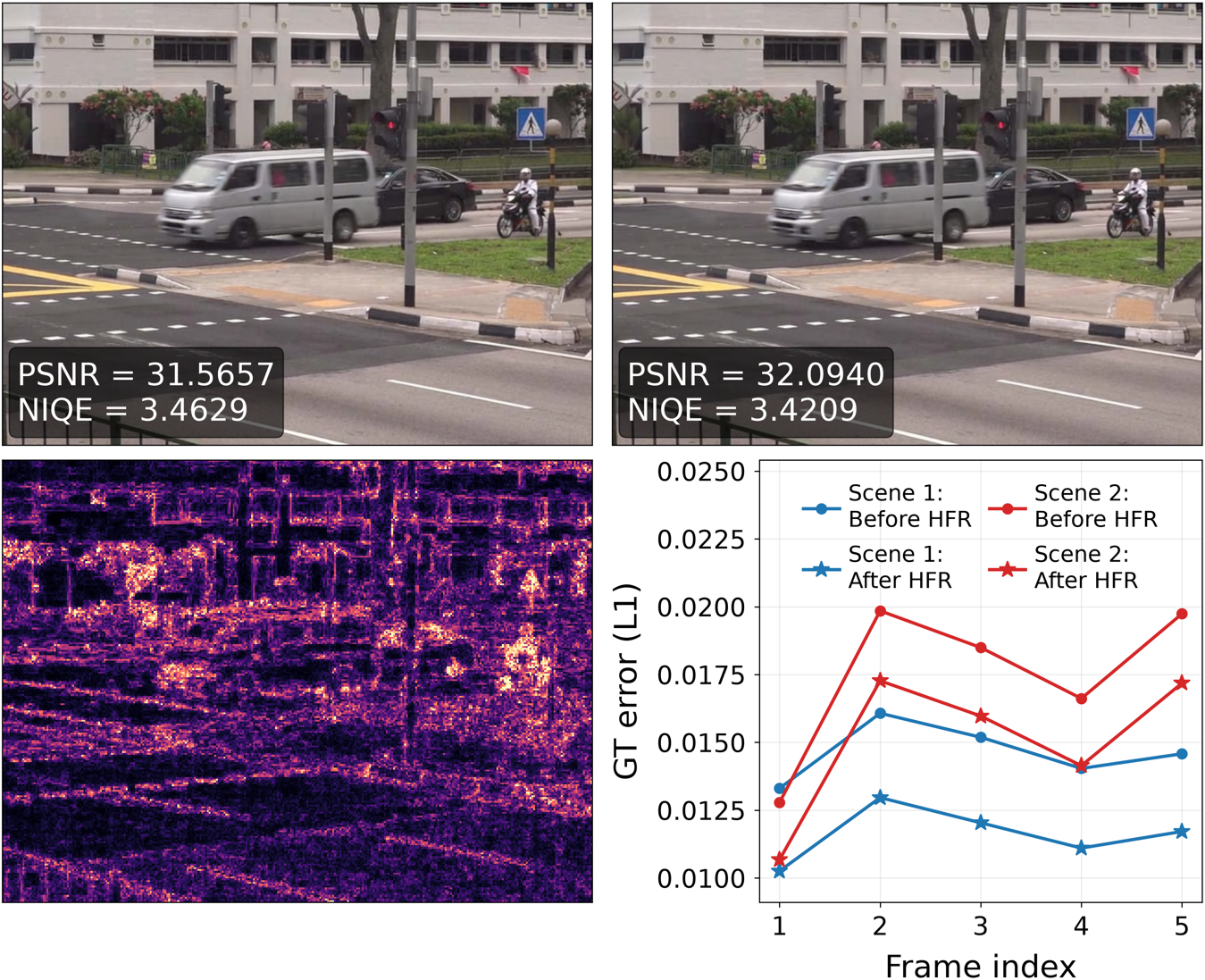}\vspace{-2mm}
\caption{HFR diagnostic. Top: outputs before and after HFR with configuration-level scores. Bottom: absolute high-band coefficient changes and per-frame RGB $L_1$ errors to GT over five frames in two scenes. Circles/stars indicate before/after HFR.}\vspace{-2mm}
\label{fig:ablation_hfr}
\end{figure}

\paragraph{High-frequency Refinement.}
Figure~\ref{fig:ablation_hfr} compares the decoded output $\hat Y^{\rm dec}$ and refined output $\hat Y$ from the same trained model.
The high-band changes concentrate around edges, text, and other fine structures, while the low-frequency wavelet coefficients remain unchanged by construction (Eq.~\ref{eq:hfr}).
HFR lowers the RGB $L_1$ error in each of the five displayed frames for both scenes.
These examples illustrate the effect of bounded high-frequency correction after diffusion decoding.
% \paragraph{High-frequency Refinement.}
% Figure~\ref{fig:ablation_hfr} compares $\hat Y^{\rm dec}$ and $\hat Y$ from the same model. The high-band difference concentrates around edges, text, and other fine structures, while the decoded low band remains unchanged by construction (Eq.~\ref{eq:hfr}). This indicates that HFR corrects detail-level errors without replacing the low-frequency scene layout already recovered by the diffusion branch. HFR lowers the per-frame RGB $L_1$ error across the five displayed frames in both scenes, and the improvement is consistent with the observed high-frequency correction pattern. The result supports using the refinement head as a bounded detail correction rather than as a second full-image reconstruction branch.

\section{Conclusion}
\label{sec:conclusion}
We presented ReCoRain, a reliability-coupled approach to event-guided video deraining.
Rather than treating correspondence, content aggregation, and noise construction separately, the framework uses a common observation assessment to regulate aligned content corrections and motion-linked noise sharing.
This initialization preserves standard-normal coordinate marginals, introduces observation-conditioned correlations, and is used in flow-matching training and sampling.
Clean-video relation supervision and bounded high-frequency refinement complement it at the output.
Full-resolution restoration and fixed-checkpoint diagnostics support coupling reconstruction evidence with temporal stochastic organization.
The approach remains dependent on estimated motion and a finite-interval trajectory approximation; extending correspondence assessment to severe occlusion and longer temporal spans is a natural direction for further study.
% Mandatory LLM Usage Statement (CVPR 2027). It does not count toward the
% eight-page limit and must stay anonymous.
%
% TODO (source-only note, prints nothing): make sure this text matches what you
% actually used, and keep it anonymous.
\section*{LLM Usage Statement}
We used ChatGPT (OpenAI) to polish selected manuscript passages, including the Related Work section, and to generate an initial draft of the framework illustration, which was subsequently revised manually. We also used it to edit selected parts of the implementation code and refine the layout of several figures and tables.
{
    \small
    \bibliographystyle{ieeenat_fullname}
    \bibliography{references_wanrain}
}

\end{document}